\documentclass[10pt, a4paper]{article}

\usepackage{lrec-coling2024} 

\title{Can We Identify Stance Without Target Arguments? A Study for Rumour Stance Classification}

\name{Yue Li, Carolina Scarton} 

\address{University of Sheffield\\
         211 Portobello, Sheffield, UK\\
         {yli381, c.scarton}@sheffield.ac.uk\\}

\abstract{
Considering a conversation thread, rumour stance classification aims to identify the opinion (e.g. agree or disagree) of replies towards a \textit{target} (rumour story). Although the target is expected to be an essential component in traditional stance classification, we show that rumour stance classification datasets contain a considerable amount of real-world data whose stance could be naturally inferred directly from the replies, 
contributing to the strong performance of the supervised models without awareness of the target. We find that current target-aware models 
underperform in cases where the context of the target is crucial. Finally, we propose a simple yet effective framework to enhance reasoning with the targets, achieving 
state-of-the-art performance on two benchmark datasets.
 \\ \newline  }

\begin{document}

\maketitleabstract

\section{Introduction}

Automatic stance classification that aims to identify the type of an expressed opinion towards a single or multiple \textit{targets}, plays a key role in many Natural Language Processing (NLP) applications, such as 
rumour analysis \citep{zubiaga2016analysing}. 
A target could be a person, an organisation, or rumour story, depending on the use case \citep{hossain-etal-2020-covidlies,zubiaga2016analysing,ferreira-vlachos-2016-emergent,allaway-mckeown-2020-zero}. The target plays a fundamental role in stance classification, being expected to appear either explicitly or implicitly, making it a key difference from sentiment analysis that can be framed as target-independent \citep{kuccuk2020stance,liu-etal-2022-target}.

Previous work shows that a BERT-based model, without awareness of the target, achieves comparable or even better performance than target-aware models on many stance classification datasets, due to spurious sentiment- and lexicon-stance correlations in the training sets \citep{kaushal-etal-2021-twt}. Similar results are observed in other context-dependent tasks, such as Natural Language Inference and Argument Reasoning
Comprehension, where models without background knowledge achieve an impressive performance due to spurious or superficial cues in the datasets \citep{poliak-etal-2018-hypothesis,niven-kao-2019-probing}. 

\begin{figure}[t!]
 \centering

 \includegraphics[width=0.35\textwidth]{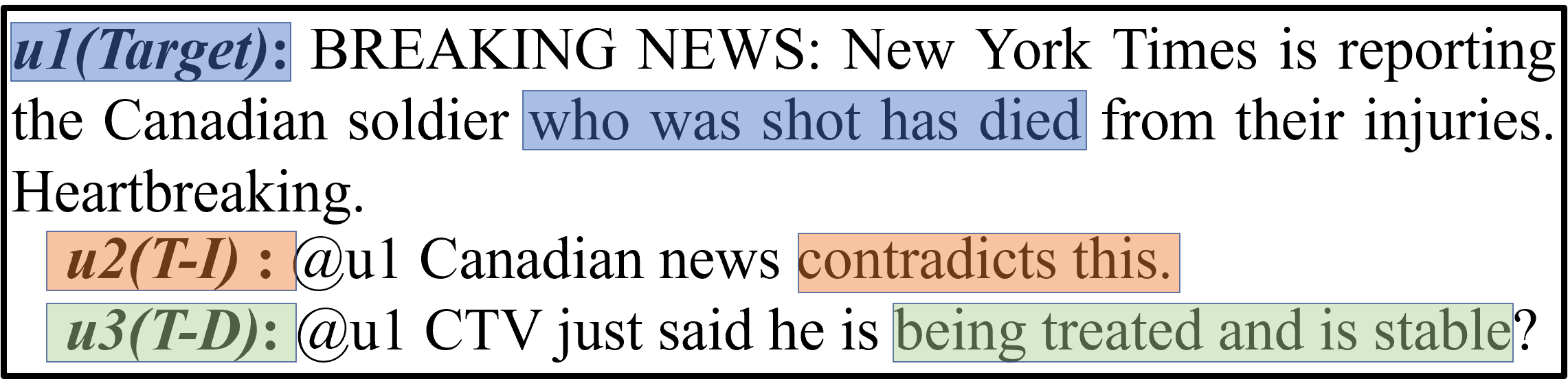}
 
 \caption{Example of Target-Independent (T-I) and Target-Dependent (T-D) direct replies that \textit{deny} a target from \citet{gorrell-etal-2019-semeval}.}
 \label{fig:example}

\end{figure}

In this paper, we further analyse the above phenomenon for \textit{rumour stance classification} on Twitter. Given a conversation initialised by a rumourous \textit{source tweet}, this task aims to classify the stance of each reply towards the rumour into \textit{support}, \textit{deny}, \textit{query} and \textit{comment}. The vagueness and lack of specificity in the reply tweets result in the disparity between rumour stance classification and traditional stance classification datasets. 
For instance, in Figure \ref{fig:example}, one can reasonably deduce that the reply from \textit{u2} disagrees with the target before reading the content of the target. This is in contrast to traditional stance classification where the stance may vary for different targets, making it always essential to consider them (e.g., \citealp{sobhani-etal-2017-dataset,conforti-etal-2020-will}). 

We empirically show that the strong behaviour of models without awareness of the target (dubbed \textit{target-oblivious}) could be explained by the existence of the reply posts whose stance can be naturally inferred without knowing the target. More importantly, we demonstrate that current state-of-the-art target-aware models lack reasoning with the target, performing unexpectedly poorly on the cases when the target is necessary. Based on our observations, we propose a simple yet effective framework which would benefit from the target-oblivious model and would also enhance the reasoning with the targets.

\section{The Role of Target Arguments} 

We conduct an annotation study by categorising the replies into \textit{target-dependent} (i.e. target is essential for stance inference) and \textit{target-independent} (i.e. target is unnecessary for stance inference). We then evaluate various models trained with or without awareness of the target (i.e. \textit{target-aware} and \textit{target-oblivious} models). 

\subsection{Data Annotation} 

\paragraph{Dataset} Three established English datasets are available for rumour stance classification on social media: \textit{PHEME} \citep{zubiaga2016analysing}, \textit{RumourEval 2017} \citep{derczynski-etal-2017-semeval} and \textit{RumourEval 2019} \citep{gorrell-etal-2019-semeval}. RumourEval 2017 consists of the English PHEME dataset, and RumourEval 2019 is an extension of the 2017 dataset. Therefore, we consider the largest RumourEval 2019 dataset.\footnote{The dataset contains Twitter and Redddit. To alleviate the impact of text length, we focus on the Twitter data only}
RumourEval 2019 training and validation sets consist of conversations regarding rumour stories which emerged during breaking news (e.g., Germanwings plane crash, and shooting in Ottawa), and the test data contains unseen rumours about natural disasters. 
The target of the stance, rumour story, is implied by the source tweet that initialises the conversation. Hence we consider the source tweet as the target.
Among the four stances, support and deny classes are the most informative for rumour verification, while the comment class is the least useful \citep{scarton-etal-2020-measuring}. Therefore, we annotate all the replies in support, deny and query classes in the validation and test sets, with 50 randomly sampled comments from each set.

\paragraph{Annotation Process} Two annotators manually categorised each reply into either target-independent or -dependent, by answering one question: “\textit{do you think you need the source tweet to infer the stance of this reply?}” Aiming to validate the annotations, annotators were also asked to classify the stance of the tweets. We then compared their assigned class with the gold standard label and, if they differed, we altered their annotation from target-independent to -dependent. Annotators did not have access to the source tweet and the tweets from validation and test sets were shuffled before annotation. The inter-annotator agreement is of 72.5\% and Cohen’s Kappa is 0.565.

\paragraph{Result} We observe a significant amount of data whose stance can be deduced without knowing the specific rumour story (Table \ref{tab:results-annotate-target}), especially in the deny and query classes. More than 50\% denies are target-independent in the validation and test sets.
Target-independent denies are tweets that directly cast doubt with negation words (e.g. “Fake news”, "This is false"). The queries tend to be target-independent, since most of them are interrogative sentences asking for more evidence. 
However, the annotators did not identify many of them due to the ambiguity or non-informativeness of the texts (e.g., “blood clot?”, “WHAT?”). Most of the target-independent supports are retweets and quote tweets, whose context is self-contained. Tweets in the comment class are less relevant to the veracity of the rumour story, however, determining their relevance normally necessitates reasoning with the rumour story itself. 

\begin{table}[t!]
\centering
\scalebox{0.65}{
\begin{tabular}{lcccc}
\hline
Dataset & Support & Deny & Query & Comment\\
\hline
Validation & 20 (29\%) & 35 (51\%) & 42 (40\%) & 0 (0\%)\\
\hline
Test & 12 (13\%) & 66 (72\%) & 17 (30\%) & 0 (0\%)\\
\hline
\end{tabular}
}
\caption{Number of target-independent tweets in each stance in the validation and test sets (proportion in brackets).} 
\label{tab:results-annotate-target}
\end{table}

\subsection{Model Evaluation} 

Given a source tweet ($s_i$), reply tweet to classify ($r_i$), other replies in the conversation ($o_i$) and stance label ($l_i$), we consider two types of supervised models: target-oblivious ($f(r_i)\rightarrow l_i$) and target-aware ($f(s_i,r_i)$ or $f(s_i,r_i,o_i)\rightarrow l_i$) models. We also evaluate a recent large language model (LLM) in zero-shot setting.

\begin{table*}[ht!]
\centering
\scalebox{0.65}{
\begin{tabular}{ll|l|lll|lll}
\hline
Type & Model & Full set & \multicolumn{3}{l}{Target-dependent subset} & \multicolumn{3}{|l}{Target-independent subset} \\ 
& & $wF_2$ & $wF_2$ & $F_2(S)$ & $F_2(D)$ & $wF2$ & $F_2(S)$ & $F_2(D)$ \\ \hline \hline
Target-oblivious & BERTweet & \textbf{0.477} & \textbf{0.346} & 0.294 & 0.206 & \textbf{0.749} & 0.615 & \textbf{0.894}\\ \hline \hline
Target-aware & BERTweet & 0.435 & 0.329 & 0.313 & 0.167 & 0.635 & 0.464 & 0.778 \\ 
& BLCU-NLP & 0.371 & 0.223 & 0.080 & 0.217 & 0.399 & 0.000 & 0.737 \\
& BUT-FIT & 0.309 & 0.176 & 0.020 & 0.047 & 0.371 & 0.102 & 0.495 \\
& Branch-LSTM & 0.150 & 0.139 & 0.020 & 0.048 & 0.142 & 0.102 & 0.056 \\
& Hierarchical-BERT & 0.235 & 0.137 & 0.065 & 0.017 & 0.234 & 0.017 & 0.293\\ \hline \hline
LLMs & LLaMA (reply) & 0.256 & 0.227 & \textbf{0.390} & 0.000 & 0.319 & 0.417 & 0.093 \\
& LLaMA (source \& reply) & 0.419 & 0.318 & 0.326 & \textbf{0.234} & 0.685 & \textbf{0.678} & 0.714 \\ \hline \hline

\end{tabular}
}
\caption[]{Model performance over the full test set, target-dependent and -independent direct replies (averaged over experiments.). $F_2(S)$ and $F_2(D)$ denote the $F_2$ scores over support and deny classes, respectively. Highest performance is in bold, with statistical significance (t test, p value<0.05).}
\label{tab:model performance}
\end{table*}

\begin{table*}[t!]
\centering
\scalebox{0.65}{
\begin{tabular}{l|cccc|cccc}
\hline
& \multicolumn{4}{c}{Target-dependent subset} & \multicolumn{4}{|c}{Target-independent subset} \\
& Support & Deny & Query & Comment & Support & Deny & Query & Comment\\
\hline
Mask Source Tweet & 40.3 & 69.9 & 98.7 & 85.7 & 43.0 & 90.8 & 98.7 & 89.0\\
\hline
Shuffle Source Tweet & 54.1 & 82.9 & 93.6 & 90.9 & 57.7 & 94.9 & 97.3 & 83.1 \\
\hline
\end{tabular}
}
\caption{The proportion (\%) of target-aware BERTweet predictions of direct replies in each class that are not influenced by the masking or shuffling of the source tweets.}
\label{tab:results-perturbation}
\end{table*}

\subsubsection{Experimental Setups}

\paragraph{Target-oblivious Models} We fine-tune different transformer-based models, whose input is the reply tweet ($f(r_i)$). 
We present the results using BERTweet \citep{nguyen-etal-2020-bertweet} (experiments with BERT \citep{devlin-etal-2019-bert} and Roberta \citep{liu2019roberta} achieved similar performance).

\paragraph{Target-aware Models} We fine-tune BERTweet, which takes input as both source and reply tweets ($f(s_i,r_i)$). We also evaluate four competitive systems that model the whole conversation thread ($f(s_i,r_i,o_i)$):\footnote{Performances of these models are lower than the figures reported in their original paper. The reason is that we do not consider the stance of the source tweet towards rumour, mainly belonging to the support class.} (1) The winner of the RumourEval 2019 shared task, i.e. \textit{BLCU-NLP} \citep{yang-etal-2019-blcu}; (2) \textit{BUT-FIT} \citep{fajcik-etal-2019-fit}, the second place in the 2019 shared task; (3) \textit{Hierarchical-BERT} \citep{yu-etal-2020-coupled}, achieving state-of-the-art performance \citep{hardalov-etal-2022-survey} on the RumourEval 2017 dataset \citep{derczynski-etal-2017-semeval}; (4) \textit{Branch-LSTM} \citep{kochkina-etal-2017-turing}, the winner of the RumourEval 2017 shared task and the baseline model for the 2019 task. 

\paragraph{LLMs} 
We experiment with the OpenAssistant LLaMA-Based Model \citep{kopf2023openassistant}.\footnote{https://huggingface.co/OpenAssistant/oasst-sft-6-llama-30b-xor} We compare the performance between when the source tweet is provided and when it is not (\textit{LLaMA (source + reply)} or \textit{LLaMA (reply)}).\footnote{Due to ethical considerations regarding the exposure of personal data (e.g., to ChatGPT), we opt to use an open-source LLM which was downloaded and hosted on our own server.} 

\paragraph{Evaluation} 
We adopt the weighted $F_2$ score proposed by \citet{scarton-etal-2020-measuring}, which gives higher weights to the support and deny classes, being more adequate to rumour stance classification. 

\subsubsection{Results}

As shown in Table \ref{tab:model performance}, not surprisingly, all the models achieve better results on the target-independent samples, since they normally contain explicit stance-associated words or signals, especially for the deny and query classes. The target-oblivious model exhibits strong performance over target-independent tweets, indicating that its performance can be attributed to the existence of these samples in the dataset. 

We expected that target-aware models, especially the ones that consider the whole conversation information, would perform significantly better than target-oblivious models on the target-dependent tweets for which the context of the source tweet is essential. However, 
BUT-FIT, Branch-LSTM and Hierarchical-BERT could not correctly predict any target-dependent supports or denies that the simple target-oblivious BERTweet fails to identify, casting doubt on the usefulness of these approaches. BLCU-NLP is the only conversation-based system that outperforms the target-oblivious model over the target-dependent denies, likely due to their data augmentation for this class. But its performance over the target-dependent supports is rather disappointing.

Target-aware BERTweet shows strength on detecting target-dependent supports, when compared with its target-oblivious counterpart; however, it falls behind on the deny class. The existence of negation words (e.g., “not”) in the target-dependent denies may contribute to the good generalisation of target-oblivious BERTweet. 

LLaMA exhibits competitive results, achieving best performance on the target-dependent samples in the support and deny classes. However, gaps still exist between the fine-tuned BERTweet models on the full test set. Without the source tweet, the performance drops significantly, except for the target-dependent supports.

\subsubsection{Target perturbations} Aiming to further investigate the role of the target in target-aware models, we experiment with two perturbations during inference: (1) Masking: the entire source tweet is replaced by a white space; (2) Shuffling: the original source tweet is replaced by a source tweet related to another rumour story so that the reply and “new” source tweets are mismatched. Both approaches should significantly change the model performance over the target-dependent tweets, provided the source tweet is properly reasoned with. We expect the comment class to be less impacted because the irrelevance between source and reply tweets should be considered as comment. We discuss the results of the target-aware BERTweet, since it is the best performing model in this category (other models showed similar results).

Masking or shuffling the source tweets has minimum impact over the predictions for the \textit{deny}, \textit{query} and \textit{comment} classes (Table \ref{tab:results-perturbation}). More than 69\% of predictions in each class stay the same, no matter whether the target is essential or not. For the \textit{support} class in which target-aware BERTweet achieves better results over target-dependent samples, 40\% to 60\% of predictions do not change. The results suggest that target-aware models may be overfitting towards the replies, behaving like a target-oblivious model.

\section{Ensemble-based Framework}

Equipped with the observation of target-independent cases and the lack of reasoning with the target in target-aware models, we propose a simple yet effective ensemble-based framework to leverage the advantage of the target-oblivious model meanwhile improving the performance over the target-dependent samples. 

We assume a pre-trained target-oblivious model ($f(r_i;\theta)=p_i$). The aim is to adopt an ensemble with a target-aware model ($f'(s_i,r_i;\theta')=q_i$) where $p_i$ and $q_i$ are posterior probability distribution over the four stance classes for a sample $i$ with a pair of source ($s_i$) and reply ($r_i$) tweets. To encourage the target-aware model to learn from target-dependent samples during training, we propose a cross-attention based architecture with a sample re-weight mechanism. 

\paragraph{Siamese Network with Cross-attention} We utilise a siamese pre-trained transformer-based network \citep{reimers-gurevych-2019-sentence} to encode the source ($s_i$) and reply ($r_i$) tweets. Then, to explicitly indicate the importance of the tokens in the reply representation ($h_{r_i}$) with respect to the source representation ($h_{s_i}$), we calculate the cross-attention \citep{vaswani2017attention} between them, with $h_{s_i}$ as the key and value, and $h_{r_i}$ as the query.  

\paragraph{Sample Re-weight} We train the model on weighted data, where the weight of instance $i$ is $1-p_{y_i}$ ($p_{y_i}$ is the posterior probability assigned to the true label $y_i$) \citep{clark-etal-2019-dont}. The intuition is to encourage the target-aware model to focus on potential target-dependent examples that the target-oblivious model gets wrong. 

\paragraph{Implementation} Target-oblivious and -aware models are based on BERTweet but our method can be easily generalised to other pre-trained language models. The optimal target-oblivious model is chosen based on the validation set. 

\subsection{Experimental Setup}

\paragraph{Datasets} We validate our proposed framework on two benchmark datasets: RumourEval 2017 and 2019 datasets.

\paragraph{Comparing Baselines} We compare with the Pretext Task-based Hierarchical Contrastive Learning model (\textit{PT-HCL}) \citep{liang2022zero}. To the best of our knowledge, PT-HCL is the only study that exploits “target-invariant/-specific features” \citep{liang2022zero} in traditional stance classification. We also present ablations for our proposed method, by removing the sample re-weighting mechanism (\textit{w/o weight}), replacing the cross-attention by self-attention on the concatenation of the source and reply tweet representations (\textit{w/o cross-att}), or both simultaneously (\textit{w/o weight,cross-att}). Performance over RumourEval 2019 dataset is comparable with models in Table \ref{tab:results-proposed}. As for RumourEval 2017, we also compare with its state-of-the-art model (Hierarchical-BERT), target-oblivious and -aware BERTweet and OpenAssistant LLaMa.

\subsection{Results}

As shown in Table \ref{tab:results-proposed}, our proposed approach outperforms PT-HCL on both datasets, also surpassing other models. Removing sample weights or cross-attention would reduce the model performance, indicating their contribution. 

\begin{table}[ht!]
\centering
\scalebox{0.65}{
\begin{tabular}{l|cc}
\hline
Method & RumourEval2019 & RumourEval2017 \\
\hline
PT-HCL & 0.452 & 0.431 \\
\hline
Hierarchical-BERT & 0.235 & 0.275\\
LLaMA & 0.419 & 0.314\\
Target-oblivious BERTweet & 0.477 & 0.425 \\
Target-aware BERTweet & 0.435 & 0.426 \\
\hline
\textbf{Proposed Method} & \textbf{0.510} & \textbf{0.452} \\
w/o weight & 0.458 & 0.421 \\
w/o cross-att & 0.438 & 0.417\\
w/o weight,cross-att & 0.451 & 0.419\\

\hline
\end{tabular} 
}
\caption{Averaged $wF_2$ over experiments for two datasets. Highest performance is in bold, with statistical significance between the proposed method (t test, p value <0.05).}
\label{tab:results-proposed}
\end{table}

We also evaluate our proposed method on target-dependent and -independent subsets. 
Comparing with Table \ref{tab:model performance}, it achieves the best results on both target-dependent ($wF_2=0.396$, with $F_2(S)=0.399$, $F_2(D)=0.211$) and -independent examples ($wF_2=0.802$, with $F_2(S)=0.732$, $F_2(D)=0.901$) , confirming that our proposed framework could not only benefit from the target-oblivious model but also enhance the inference between source and reply tweets. 


\section{Conclusion}

In this paper, we explore the role of the target in rumour stance classification. Our study suggests the strong performance of the target-oblivious model could be explained by the existence of target-independent texts in real-world data. We point out the unexpected weakness of the target-aware models and consequently propose a cross-attention based architecture with sample re-weight mechanism, achieving the best result on two benchmark datasets. We also release our annotation to facilitate future research and model evaluations.\footnote{The link will be available upon acceptance.}

\section{Acknowledgements}


This work is funded by the European Union under action number 2020-EU-IA-0282 and agreement number INEA/CEF/ICT/A2020/2381686 (EDMO Ireland).\footnote{\url{https://edmohub.ie}} and by EMIF managed by the Calouste Gulbenkian Foundation\footnote{The sole responsibility for any content supported by the European Media and Information Fund lies with the author(s) and it may not necessarily reflect the positions of the EMIF and the Fund Partners, the Calouste Gulbenkian Foundation and the European University Institute.} under the "Supporting Research into Media, Disinformation and Information Literacy Across Europe" call (ExU -- project number: 291191).\footnote{\url{exuproject.sites.sheffield.ac.uk}} Yue Li is supported by a Sheffield–China Scholarships Council PhD Scholarship.

\section{Bibliographical References}\label{sec:reference}

\bibliographystyle{lrec-coling2024-natbib}
\bibliography{lrec-coling2024-example}

\begin{thebibliography}{27}
\expandafter\ifx\csname natexlab\endcsname\relax\def\natexlab#1{#1}\fi

\bibitem[{Allaway and McKeown(2020)}]{allaway-mckeown-2020-zero}
Emily Allaway and Kathleen McKeown. 2020.
\newblock \href {https://doi.org/10.18653/v1/2020.emnlp-main.717} {{Z}ero-{S}hot {S}tance {D}etection: {A} {D}ataset and {M}odel using {G}eneralized {T}opic {R}epresentations}.
\newblock In \emph{Proceedings of the 2020 Conference on Empirical Methods in Natural Language Processing (EMNLP)}, pages 8913--8931, Online. Association for Computational Linguistics.

\bibitem[{Clark et~al.(2019)Clark, Yatskar, and Zettlemoyer}]{clark-etal-2019-dont}
Christopher Clark, Mark Yatskar, and Luke Zettlemoyer. 2019.
\newblock \href {https://doi.org/10.18653/v1/D19-1418} {Don{'}t take the easy way out: Ensemble based methods for avoiding known dataset biases}.
\newblock In \emph{Proceedings of the 2019 Conference on Empirical Methods in Natural Language Processing and the 9th International Joint Conference on Natural Language Processing (EMNLP-IJCNLP)}, pages 4069--4082, Hong Kong, China. Association for Computational Linguistics.

\bibitem[{Conforti et~al.(2020)Conforti, Berndt, Pilehvar, Giannitsarou, Toxvaerd, and Collier}]{conforti-etal-2020-will}
Costanza Conforti, Jakob Berndt, Mohammad~Taher Pilehvar, Chryssi Giannitsarou, Flavio Toxvaerd, and Nigel Collier. 2020.
\newblock \href {https://doi.org/10.18653/v1/2020.acl-main.157} {Will-they-won{'}t-they: A very large dataset for stance detection on {T}witter}.
\newblock In \emph{Proceedings of the 58th Annual Meeting of the Association for Computational Linguistics}, pages 1715--1724, Online. Association for Computational Linguistics.

\bibitem[{Derczynski et~al.(2017)Derczynski, Bontcheva, Liakata, Procter, Wong Sak~Hoi, and Zubiaga}]{derczynski-etal-2017-semeval}
Leon Derczynski, Kalina Bontcheva, Maria Liakata, Rob Procter, Geraldine Wong Sak~Hoi, and Arkaitz Zubiaga. 2017.
\newblock \href {https://doi.org/10.18653/v1/S17-2006} {{S}em{E}val-2017 task 8: {R}umour{E}val: Determining rumour veracity and support for rumours}.
\newblock In \emph{Proceedings of the 11th International Workshop on Semantic Evaluation ({S}em{E}val-2017)}, pages 69--76, Vancouver, Canada. Association for Computational Linguistics.

\bibitem[{Devlin et~al.(2019)Devlin, Chang, Lee, and Toutanova}]{devlin-etal-2019-bert}
Jacob Devlin, Ming-Wei Chang, Kenton Lee, and Kristina Toutanova. 2019.
\newblock \href {https://doi.org/10.18653/v1/N19-1423} {{BERT}: Pre-training of deep bidirectional transformers for language understanding}.
\newblock In \emph{Proceedings of the 2019 Conference of the North {A}merican Chapter of the Association for Computational Linguistics: Human Language Technologies, Volume 1 (Long and Short Papers)}, pages 4171--4186, Minneapolis, Minnesota. Association for Computational Linguistics.

\bibitem[{Fajcik et~al.(2019)Fajcik, Smrz, and Burget}]{fajcik-etal-2019-fit}
Martin Fajcik, Pavel Smrz, and Lukas Burget. 2019.
\newblock \href {https://doi.org/10.18653/v1/S19-2192} {{BUT}-{FIT} at {S}em{E}val-2019 task 7: Determining the rumour stance with pre-trained deep bidirectional transformers}.
\newblock In \emph{Proceedings of the 13th International Workshop on Semantic Evaluation}, pages 1097--1104, Minneapolis, Minnesota, USA. Association for Computational Linguistics.

\bibitem[{Ferreira and Vlachos(2016)}]{ferreira-vlachos-2016-emergent}
William Ferreira and Andreas Vlachos. 2016.
\newblock \href {https://doi.org/10.18653/v1/N16-1138} {{E}mergent: a novel data-set for stance classification}.
\newblock In \emph{Proceedings of the 2016 Conference of the North {A}merican Chapter of the Association for Computational Linguistics: Human Language Technologies}, pages 1163--1168, San Diego, California. Association for Computational Linguistics.

\bibitem[{Gorrell et~al.(2019)Gorrell, Kochkina, Liakata, Aker, Zubiaga, Bontcheva, and Derczynski}]{gorrell-etal-2019-semeval}
Genevieve Gorrell, Elena Kochkina, Maria Liakata, Ahmet Aker, Arkaitz Zubiaga, Kalina Bontcheva, and Leon Derczynski. 2019.
\newblock \href {https://doi.org/10.18653/v1/S19-2147} {{S}em{E}val-2019 task 7: {R}umour{E}val, determining rumour veracity and support for rumours}.
\newblock In \emph{Proceedings of the 13th International Workshop on Semantic Evaluation}, pages 845--854, Minneapolis, Minnesota, USA. Association for Computational Linguistics.

\bibitem[{Hardalov et~al.(2022)Hardalov, Arora, Nakov, and Augenstein}]{hardalov-etal-2022-survey}
Momchil Hardalov, Arnav Arora, Preslav Nakov, and Isabelle Augenstein. 2022.
\newblock \href {https://doi.org/10.18653/v1/2022.findings-naacl.94} {A survey on stance detection for mis- and disinformation identification}.
\newblock In \emph{Findings of the Association for Computational Linguistics: NAACL 2022}, pages 1259--1277, Seattle, United States. Association for Computational Linguistics.

\bibitem[{Hossain et~al.(2020)Hossain, Logan~IV, Ugarte, Matsubara, Young, and Singh}]{hossain-etal-2020-covidlies}
Tamanna Hossain, Robert~L. Logan~IV, Arjuna Ugarte, Yoshitomo Matsubara, Sean Young, and Sameer Singh. 2020.
\newblock \href {https://doi.org/10.18653/v1/2020.nlpcovid19-2.11} {{COVIDL}ies: Detecting {COVID}-19 misinformation on social media}.
\newblock In \emph{Proceedings of the 1st Workshop on {NLP} for {COVID}-19 (Part 2) at {EMNLP} 2020}, Online. Association for Computational Linguistics.

\bibitem[{Kaushal et~al.(2021)Kaushal, Saha, and Ganguly}]{kaushal-etal-2021-twt}
Ayush Kaushal, Avirup Saha, and Niloy Ganguly. 2021.
\newblock \href {https://doi.org/10.18653/v1/2021.naacl-main.303} {t{WT}{--}{WT}: A dataset to assert the role of target entities for detecting stance of tweets}.
\newblock In \emph{Proceedings of the 2021 Conference of the North American Chapter of the Association for Computational Linguistics: Human Language Technologies}, pages 3879--3889, Online. Association for Computational Linguistics.

\bibitem[{Kochkina et~al.(2017)Kochkina, Liakata, and Augenstein}]{kochkina-etal-2017-turing}
Elena Kochkina, Maria Liakata, and Isabelle Augenstein. 2017.
\newblock \href {https://doi.org/10.18653/v1/S17-2083} {{T}uring at {S}em{E}val-2017 task 8: Sequential approach to rumour stance classification with branch-{LSTM}}.
\newblock In \emph{Proceedings of the 11th International Workshop on Semantic Evaluation ({S}em{E}val-2017)}, pages 475--480, Vancouver, Canada. Association for Computational Linguistics.

\bibitem[{K{\"o}pf et~al.(2023)K{\"o}pf, Kilcher, von R{\"u}tte, Anagnostidis, Tam, Stevens, Barhoum, Duc, Stanley, Nagyfi et~al.}]{kopf2023openassistant}
Andreas K{\"o}pf, Yannic Kilcher, Dimitri von R{\"u}tte, Sotiris Anagnostidis, Zhi-Rui Tam, Keith Stevens, Abdullah Barhoum, Nguyen~Minh Duc, Oliver Stanley, Rich{\'a}rd Nagyfi, et~al. 2023.
\newblock Openassistant conversations--democratizing large language model alignment.
\newblock \emph{arXiv preprint arXiv:2304.07327}.

\bibitem[{K{\"u}{\c{c}}{\"u}k and Can(2020)}]{kuccuk2020stance}
Dilek K{\"u}{\c{c}}{\"u}k and Fazli Can. 2020.
\newblock \href {https://doi.org/10.1145/3369026} {Stance detection: A survey}.
\newblock \emph{ACM Computing Surveys (CSUR)}, 53(1):1--37.

\bibitem[{Liang et~al.(2022)Liang, Chen, Gui, He, Yang, and Xu}]{liang2022zero}
Bin Liang, Zixiao Chen, Lin Gui, Yulan He, Min Yang, and Ruifeng Xu. 2022.
\newblock \href {https://doi.org/10.1145/3485447.3511994} {Zero-shot stance detection via contrastive learning}.
\newblock In \emph{Proceedings of the ACM Web Conference 2022}, pages 2738--2747.

\bibitem[{Liu et~al.(2022)Liu, Lin, Ji, Li, Fu, and Wang}]{liu-etal-2022-target}
Rui Liu, Zheng Lin, Huishan Ji, Jiangnan Li, Peng Fu, and Weiping Wang. 2022.
\newblock \href {https://aclanthology.org/2022.coling-1.605} {Target really matters: Target-aware contrastive learning and consistency regularization for few-shot stance detection}.
\newblock In \emph{Proceedings of the 29th International Conference on Computational Linguistics}, pages 6944--6954, Gyeongju, Republic of Korea. International Committee on Computational Linguistics.

\bibitem[{Liu et~al.(2019)Liu, Ott, Goyal, Du, Joshi, Chen, Levy, Lewis, Zettlemoyer, and Stoyanov}]{liu2019roberta}
Yinhan Liu, Myle Ott, Naman Goyal, Jingfei Du, Mandar Joshi, Danqi Chen, Omer Levy, Mike Lewis, Luke Zettlemoyer, and Veselin Stoyanov. 2019.
\newblock \href {https://arxiv.org/abs/1907.11692} {Roberta: A robustly optimized bert pretraining approach}.
\newblock \emph{arXiv preprint arXiv:1907.11692}.

\bibitem[{Nguyen et~al.(2020)Nguyen, Vu, and Tuan~Nguyen}]{nguyen-etal-2020-bertweet}
Dat~Quoc Nguyen, Thanh Vu, and Anh Tuan~Nguyen. 2020.
\newblock \href {https://doi.org/10.18653/v1/2020.emnlp-demos.2} {{BERT}weet: A pre-trained language model for {E}nglish tweets}.
\newblock In \emph{Proceedings of the 2020 Conference on Empirical Methods in Natural Language Processing: System Demonstrations}, pages 9--14, Online. Association for Computational Linguistics.

\bibitem[{Niven and Kao(2019)}]{niven-kao-2019-probing}
Timothy Niven and Hung-Yu Kao. 2019.
\newblock \href {https://doi.org/10.18653/v1/P19-1459} {Probing neural network comprehension of natural language arguments}.
\newblock In \emph{Proceedings of the 57th Annual Meeting of the Association for Computational Linguistics}, pages 4658--4664, Florence, Italy. Association for Computational Linguistics.

\bibitem[{Poliak et~al.(2018)Poliak, Naradowsky, Haldar, Rudinger, and Van~Durme}]{poliak-etal-2018-hypothesis}
Adam Poliak, Jason Naradowsky, Aparajita Haldar, Rachel Rudinger, and Benjamin Van~Durme. 2018.
\newblock \href {https://doi.org/10.18653/v1/S18-2023} {Hypothesis only baselines in natural language inference}.
\newblock In \emph{Proceedings of the Seventh Joint Conference on Lexical and Computational Semantics}, pages 180--191, New Orleans, Louisiana. Association for Computational Linguistics.

\bibitem[{Reimers and Gurevych(2019)}]{reimers-gurevych-2019-sentence}
Nils Reimers and Iryna Gurevych. 2019.
\newblock \href {https://doi.org/10.18653/v1/D19-1410} {Sentence-{BERT}: Sentence embeddings using {S}iamese {BERT}-networks}.
\newblock In \emph{Proceedings of the 2019 Conference on Empirical Methods in Natural Language Processing and the 9th International Joint Conference on Natural Language Processing (EMNLP-IJCNLP)}, pages 3982--3992, Hong Kong, China. Association for Computational Linguistics.

\bibitem[{Scarton et~al.(2020)Scarton, Silva, and Bontcheva}]{scarton-etal-2020-measuring}
Carolina Scarton, Diego Silva, and Kalina Bontcheva. 2020.
\newblock \href {https://aclanthology.org/2020.aacl-main.92} {Measuring what counts: The case of rumour stance classification}.
\newblock In \emph{Proceedings of the 1st Conference of the Asia-Pacific Chapter of the Association for Computational Linguistics and the 10th International Joint Conference on Natural Language Processing}, pages 925--932, Suzhou, China. Association for Computational Linguistics.

\bibitem[{Sobhani et~al.(2017)Sobhani, Inkpen, and Zhu}]{sobhani-etal-2017-dataset}
Parinaz Sobhani, Diana Inkpen, and Xiaodan Zhu. 2017.
\newblock \href {https://aclanthology.org/E17-2088} {A dataset for multi-target stance detection}.
\newblock In \emph{Proceedings of the 15th Conference of the {E}uropean Chapter of the Association for Computational Linguistics: Volume 2, Short Papers}, pages 551--557, Valencia, Spain. Association for Computational Linguistics.

\bibitem[{Vaswani et~al.(2017)Vaswani, Shazeer, Parmar, Uszkoreit, Jones, Gomez, Kaiser, and Polosukhin}]{vaswani2017attention}
Ashish Vaswani, Noam Shazeer, Niki Parmar, Jakob Uszkoreit, Llion Jones, Aidan~N Gomez, {\L}ukasz Kaiser, and Illia Polosukhin. 2017.
\newblock \href {https://proceedings.neurips.cc/paper_files/paper/2017/file/3f5ee243547dee91fbd053c1c4a845aa-Paper.pdf} {Attention is all you need}.
\newblock \emph{Advances in neural information processing systems}, 30.

\bibitem[{Yang et~al.(2019)Yang, Xie, Liu, and Yu}]{yang-etal-2019-blcu}
Ruoyao Yang, Wanying Xie, Chunhua Liu, and Dong Yu. 2019.
\newblock \href {https://doi.org/10.18653/v1/S19-2191} {{BLCU}{\_}{NLP} at {S}em{E}val-2019 task 7: An inference chain-based {GPT} model for rumour evaluation}.
\newblock In \emph{Proceedings of the 13th International Workshop on Semantic Evaluation}, pages 1090--1096, Minneapolis, Minnesota, USA. Association for Computational Linguistics.

\bibitem[{Yu et~al.(2020)Yu, Jiang, Khoo, Chieu, and Xia}]{yu-etal-2020-coupled}
Jianfei Yu, Jing Jiang, Ling Min~Serena Khoo, Hai~Leong Chieu, and Rui Xia. 2020.
\newblock \href {https://doi.org/10.18653/v1/2020.emnlp-main.108} {Coupled hierarchical transformer for stance-aware rumor verification in social media conversations}.
\newblock In \emph{Proceedings of the 2020 Conference on Empirical Methods in Natural Language Processing (EMNLP)}, pages 1392--1401, Online. Association for Computational Linguistics.

\bibitem[{Zubiaga et~al.(2016)Zubiaga, Liakata, Procter, Wong Sak~Hoi, and Tolmie}]{zubiaga2016analysing}
Arkaitz Zubiaga, Maria Liakata, Rob Procter, Geraldine Wong Sak~Hoi, and Peter Tolmie. 2016.
\newblock \href {https://doi.org/10.1371/journal.pone.0150989} {Analysing how people orient to and spread rumours in social media by looking at conversational threads}.
\newblock \emph{PloS one}, 11(3):e0150989.

\end{thebibliography}


\end{document}